\documentclass[letterpaper]{article} 
\usepackage{aaai2026}  
\usepackage{times}  
\usepackage{helvet}  
\usepackage{courier}  
\usepackage[hyphens]{url}  
\usepackage{graphicx} 
\usepackage{natbib}  
\usepackage{caption} 
\usepackage{algorithm}
\usepackage{algorithmic}
\usepackage{amsmath}
\usepackage{amssymb}
\usepackage{empheq}
\usepackage{booktabs}
\usepackage{multirow}
\usepackage{utfsym}
\usepackage{overpic}
\usepackage{threeparttable}
\usepackage{bbding}

\usepackage[hidelinks]{hyperref}

\usepackage{newfloat}
\usepackage{listings}
\DeclareCaptionStyle{ruled}{labelfont=normalfont,labelsep=colon,strut=off} 
\floatstyle{ruled}
\newfloat{listing}{tb}{lst}{}
\floatname{listing}{Listing}
\title{Beyond Boundaries: Leveraging Vision Foundation Models for Source-Free Object Detection}
\author{
    Huizai Yao\textsuperscript{\rm 1},
    Sicheng Zhao\textsuperscript{\rm 2},
    Pengteng Li\textsuperscript{\rm 1},
    Yi Cui\textsuperscript{\rm 1},
    Shuo Lu\textsuperscript{\rm 3}, \\
    Weiyu Guo\textsuperscript{\rm 1},
    Yunfan Lu\textsuperscript{\rm 1},
    Yijie Xu\textsuperscript{\rm 1},
    Hui Xiong\textsuperscript{\rm 1,\rm 4}\thanks{Corresponding Author (xionghui@ust.hk).}
}
\affiliations{
    \textsuperscript{\rm 1} Thrust of Artificial Intelligence, The Hong Kong University of Science and Technology (Guangzhou), China \\

    \textsuperscript{\rm 2} Department of Psychological and Cognitive Sciences, Tsinghua University, China \\
    \textsuperscript{\rm 3} NLPR \& MAIS, Institute of Automation, Chinese Academy of Sciences, China\\
    \textsuperscript{\rm 4} Department of Computer Science and Engineering, The Hong Kong University of Science and Technology, Hong Kong SAR, China\\
    huizai.yao@connect.hkust-gz.edu.cn
}

\usepackage{bibentry}

\begin{document}

\maketitle

\begin{abstract}

Source-Free Object Detection (SFOD) aims to adapt a source-pretrained object detector to a target domain without access to source data. However, existing SFOD methods predominantly rely on internal knowledge from the source model, which limits their capacity to generalize across domains and often results in biased pseudo-labels, thereby hindering both transferability and discriminability. In contrast, Vision Foundation Models (VFMs), pretrained on massive and diverse data, exhibit strong perception capabilities and broad generalization, yet their potential remains largely untapped in the SFOD setting. In this paper, we propose a novel SFOD framework that leverages VFMs as external knowledge sources to jointly enhance feature alignment and label quality. Specifically, we design three VFM-based modules: (1) Patch-weighted Global Feature Alignment (PGFA) distills global features from VFMs using patch-similarity-based weighting to enhance global feature transferability; (2) Prototype-based Instance Feature Alignment (PIFA) performs instance-level contrastive learning guided by momentum-updated VFM prototypes; and (3) Dual-source Enhanced Pseudo-label Fusion (DEPF) fuses predictions from detection VFMs and teacher models via an entropy-aware strategy to yield more reliable supervision. Extensive experiments on six benchmarks demonstrate that our method achieves state-of-the-art SFOD performance, validating the effectiveness of integrating VFMs to simultaneously improve transferability and discriminability.
\end{abstract}

\begin{center}
\small
\textbf{Code} — \url{https://github.com/HuizaiVictorYao/VFM_SFOD}\quad
\end{center}


\begin{figure}[!t]
\begin{center}
\includegraphics[width=\linewidth]{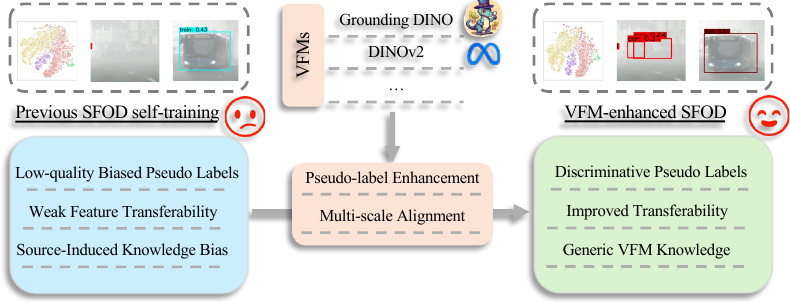}
\caption{Illustration of our VFM-enhanced SFOD motivation. Our method leverages general VFM knowledge in VFMs such as DINOv2 (visual encoder) and Grounding DINO (vision-language detector) to address pseudo-label bias and multi-scale feature misalignment, resulting in improved transferability and discriminability compared with previous self-training pipelines of SFOD.}
\label{fig:motivation}
\end{center}
\end{figure}

\section{Introduction}
Object detection, a fundamental task in computer vision that aims to accurately localize and classify objects in images or videos, has witnessed rapid advancements in the past decade. While various detectors have been extensively developed, they typically assume that training and testing data (i.e., source and target domains) are independently and identically distributed (\textit{i.i.d.}). This assumption often fails in real-world scenarios, leading to a significant drop in performance when these detectors are directly applied to unseen domains.

To alleviate this issue, Domain Adaptive Object Detection (DAOD) has been widely studied. DAOD aims to transfer a detector trained on labeled source data to unlabeled target data. However, their dependence on source data limits their applicability in scenarios involving privacy, security, or data transmission constraints~\cite{li2024comprehensive,lu2025out,zhao2023toward}. To address this, Source-Free Object Detection (SFOD) has emerged as a promising alternative, aiming to adapt a source-pretrained detector to a target domain using only unlabeled target data~\cite{vs2023instance,chu2023adversarial,yoon2024enhancing,hao2024simplifying,khanh2024dynamic,zhao2024multi}.

Previous research has identified two critical properties for effective cross-domain adaptation: transferability and discriminability~\cite{kundu2022balancing, li2024comprehensive}. Transferability ensures that feature representations remain domain-invariant, while discriminability guarantees separability between object categories. Despite this, current SFOD methods primarily focus on knowledge distillation within the teacher-student paradigm—through feature alignment~\cite{li2022source}, adversarial learning~\cite{chu2023adversarial}, or pseudo-label refinement~\cite{chen2023exploiting, zhao2024multi}, without breaking out of the internal semantic space provided by the source-pretrained detector. This closed-loop design fails to overcome the limitations of the original backbone, which often lacks semantic richness and precise decision boundaries, especially under large domain shifts. Consequently, both transferability and discriminability remain limited under domain shifts due to the restricted and homogeneous nature of internal representations in current SFOD approaches.

To address this, we turn to external knowledge sources, particularly Vision Foundation Models (VFMs). These large-scale models, pretrained on massive and diverse data, offer powerful generalization and strong visual perception. Yet, their potential remains underexplored in the SFOD setting. VFMs, including visual encoders such as DINOv2~\cite{oquab2023dinov2} and vision-language detectors such as Grounding DINO~\cite{liu2024grounding}, provide rich, transferable features and robust semantic priors, making them well-suited to augment both feature representation and label quality under source-free constraints. Some recent methods have begun exploring the use of VFMs in domain adaptation. For instance, DINO Teacher (DT)\cite{lavoie2025large} aligns detector and VFM features but requires source data and supervision, violating the SFOD assumption. CODA\cite{li2024cloud} explores external detections to enhance discriminability in cloud deployment settings, but overlooks improvements to feature-level transferability. Thus, how to fully and effectively leverage VFMs for improving SFOD in terms of discriminability and transferability remains an open and underexplored problem.

\begin{figure*}[!t]
\centering
\includegraphics[width=\linewidth]{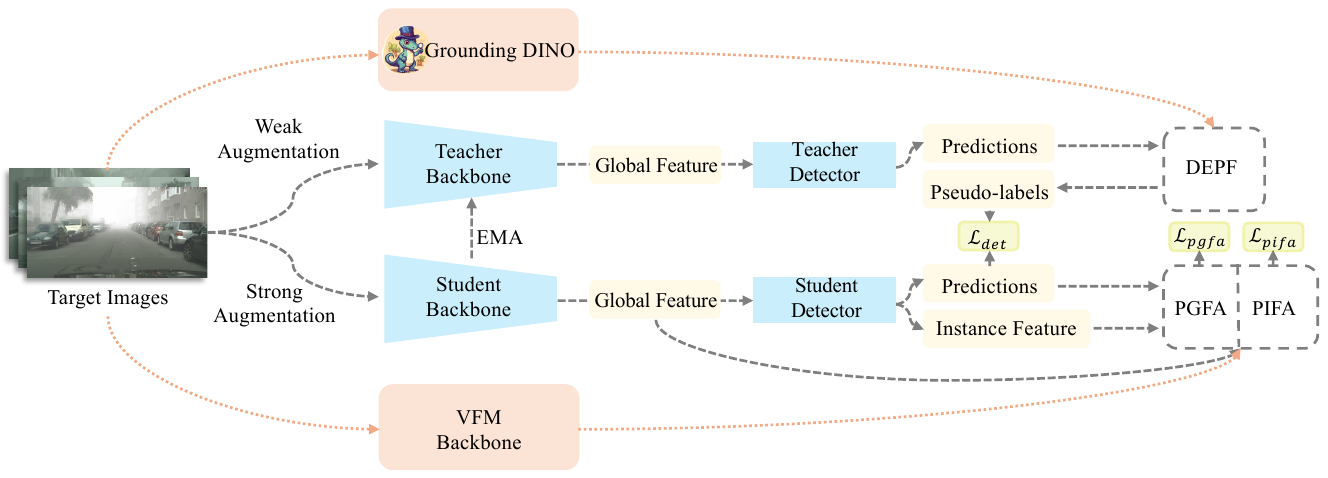}
\caption{Overview of the proposed method. Unlabeled target images are fed into teacher and student models for self-training with detection loss. DEPF fuses teacher and Grounding DINO predictions to generate refined pseudo-labels, enhancing discriminability. PGFA and PIFA align multi-scale features from the student and VFM, leveraging the generic VFM space to improve transferability.} 
\label{fig:method}
\end{figure*}

To tackle this challenge, we propose a novel VFM-enhanced SFOD framework, as illustrated in Fig.~\ref{fig:method}. Unlike previous SFOD approaches that suffer from low-quality pseudo-labels, weak transferability, and source-induced knowledge bias, our method incorporates VFMs through three key modules to address these limitations. To enhance global feature-level transferability, we propose \textbf{Patch-weighted Global Feature Alignment (PGFA)}. This module improves feature transferability by aligning the student’s feature space with that of a VFM such as DINOv2~\cite{oquab2023dinov2}, guided by patch-wise similarity weights. It encourages the student to learn from VFM’s rich, domain-agnostic representations. Considering the importance of instance features in object detection, we propose \textbf{Prototype-based Instance Feature Alignment (PIFA)}. This component constructs momentum-based prototypes from VFM features and performs contrastive learning at the instance level, enhancing fine-grained feature alignment and improving instance-level feature transferability and discriminability. To further improve pseudo-label discriminability in the self-training pipeline, we propose \textbf{Dual-source Enhanced Pseudo-label Fusion (DEPF)}. In DEPF, we fuse predictions from object detection VFMs such as Grounding DINO~\cite{liu2024grounding} and the teacher model using instance-level uncertainty-weighted box fusion. This produces cleaner, more reliable supervision signals without relying solely on the biased teacher.

As shown in Fig.~\ref{fig:motivation}, the proposed framework bridges the gap between conventional SFOD and the rich representation space of VFMs. By integrating pseudo-label enhancement, global feature alignment, and instance-level alignment, our method significantly boosts both discriminability and transferability. The resulting detector benefits from the broad generalization capability of VFMs and the structure-aware learning of the student-teacher paradigm.

Our main contributions are summarized as follows:
\begin{itemize}
    \item We identify a key limitation in current SFOD methods: the underutilization of Vision Foundation Models. To this end, we propose a unified framework that integrates VFMs to address both feature transferability and label discriminability.
    \item We design three lightweight but effective VFM-based components: \textbf{PGFA}, \textbf{PIFA}, and \textbf{DEPF}, which enhance global and instance-level alignment and improve pseudo-label quality through multi-source fusion, synergistically improve both discriminability and transferability.
    \item We conduct extensive experiments on multiple cross-domain object detection benchmarks. Results demonstrate that our method achieves state-of-the-art SFOD performance, validating the effectiveness of leveraging VFMs for cross-domain adaptation.
\end{itemize}

\section{Related Work}
\subsection{Domain Adaptive Object Detection}
Domain Adaptive Object Detection (DAOD) aims to transfer detection knowledge from labeled source domains to unlabeled target domains. Prior methods leverage self-training with teacher-student models~\cite{chen2022learning,kennerley2024cat,yao2021multi}, fine-grained alignment via graph matching~\cite{li2022sigma,li2023igg}, and data augmentation or image translation for improved generalization~\cite{liu2021domain}. More recently, Vision-Language or Foundation Models (VLMs/VFMs) have been introduced to enhance semantic robustness~\cite{li2023learning,li2024comprehensive,lavoie2025large}. However, most DAOD methods require access to source data, which conflicts with privacy constraints. In contrast, our work operates in a source-free setting, without using any labeled source images.

\subsection{Source-Free Object Detection}
SFOD (Source-Free Object Detection) aims to transfer a source pretrained model to unsupervised target domain without access to any source images. Previous SFOD approaches perform efficient knowledge transfer by techniques such as feature alignment~\cite{li2022source, vs2023instance, yao2025source} or refined pseudo-labelling~\cite{chen2023exploiting,zhang2023refined,zhao2024multi}. However, these methods operate on internal knowledge and limited decision boundary of source pretrained model, making the target performance rather limited. VG-DETR~\cite{han2025vfm} similarly employs VFMs for semi-supervised SFOD, but uses feature clustering for pseudo-labeling, leading to lower efficiency and reliability. In contrast, our method considers the rapid development of VFMs and leverages the rich semantic and broad decision boundary from VFMs pretrained by large-scale data.

\subsection{Vision Foundation Models}
VFMs are large pre‑trained models that provide a versatile foundation for vision tasks, offering broad generalization and easy adaptation via prompting or light tuning~\cite{awais2025foundation}. Recent VFMs have shown promising performance on various vision tasks such as Segment Anything Model for segmentation~\cite{kirillov2023segment}, Grounding DINO for object detection~\cite{liu2024grounding}, Depth Anything~\cite{yang2024depth} and DINOv2~\cite{oquab2023dinov2} for depth estimation. Meanwhile, VLMs and MLLMs further extend this foundation to joint multimodal understanding~\cite{lu2025uni,wu2025boosting,wang2025simple}. In this work, we leverage the rich semantic prior and strong visual perception of VFMs to effectively enhance SFOD discriminability and transferability.

\section{Method}
\subsection{Preliminaries}
\noindent \textbf{Problem Setup.} In SFOD approaches, the main goal is to transfer a  detection model $\theta_S$ from source domain $D_S$ and target domain $D_T$ with unsupervised target dataset $X_T = \{x_T^i\}_{i=1}^{N_T}$. The source model is pretrained from source dataset $X_S = \{x_S^i, y_S^i\}_{i=1}^{N_S}$ where each $(x_S^i, y_S^i)$ is \textit{i.i.d.} sampled from the source domain distribution $D_S$. Similarly, each $x_T^i$ is \textit{i.i.d.} sampled from $D_T$ but the ground-truth label is unknown. Finally, the transferred detection model $\theta_T: x_T \to y_p$ is of good cross-domain generalization and perform well on target domain.

\noindent \textbf{Mean Teacher Self-Training Pipeline.} Following previous approaches~\cite{li2022source,vs2023instance}, we adopt Mean Teacher~\cite{tarvainen2017mean} framework as the baseline. Firstly, a teacher model $\theta_{\mathrm{te}}$ and a student model $\theta_{\mathrm{st}}$ is initialized by the same parameter of the source pretrained model. In the adaptation stage, a batch of target samples is weakly and strongly augmented, and input to $\theta_{\mathrm{te}}$ and $\theta_{\mathrm{st}}$ respectively. $\theta_{\mathrm{te}}$ generates pseudo-labels as supervision signals for $\theta_{\mathrm{st}}$, and $\theta_{\mathrm{st}}$ is updated with loss backpropagation. $\theta_{\mathrm{te}}$ does not update from backpropagation but an exponential moving average (EMA) from $\theta_{\mathrm{st}}$ parameters. This progress can be denoted as:
\begin{subequations}
\begin{empheq}[left=\empheqlbrace]{align}
\theta_{\mathrm{st}} &\leftarrow \theta_{\mathrm{st}} + \eta\frac{\partial(\mathcal{L}_{\mathrm{tot}})}{\partial\theta_{\mathrm{st}}}, 
\label{eq:student_update} \\
\theta_{\mathrm{te}} &\leftarrow \alpha \theta_{\mathrm{te}} + (1-\alpha)\theta_{\mathrm{st}},
\label{eq:teacher_update}
\end{empheq}
\end{subequations}
where $\mathcal{L}_{\mathrm{tot}}$ denotes the total loss, $\eta$ denotes learning rate, and $\alpha$ denotes EMA coefficient.

\subsection{Dual-source Enhanced Pseudo-label Fusion}

Starting from the mean teacher baseline, a central challenge in SFOD lies in generating reliable pseudo-labels, which directly affect the model’s discriminability. Existing approaches attempt to refine low-confidence predictions or leverage soft-label learning~\cite{yoon2024enhancing, chen2023exploiting}, but they inherently rely on the biased knowledge of the source model, often leading to overconfident or inaccurate labels. DT~\cite{lavoie2025large} circumvents this by training a DINOv2-based labeler on source data to generate pseudo-labels for the target domain. However, this violates the core source-free constraint of SFOD and proves ineffective when adapted to noisy target data. To address this, we explore the use of powerful object detection VFMs such as Grounding DINO~\cite{liu2024grounding}, which can act as zero-shot labelers in the target domain. However, while VFMs offer strong generalization, they often underperform under domain shifts~\cite{zhang2024improving}. Motivated by this, we propose \textbf{Dual-source Enhanced Pseudo-label Fusion (DEPF)}, which fuses the generalizable predictions from VFMs with the domain-adapted cues from the teacher model. This is achieved via an entropy-guided fusion strategy that dynamically balances contributions from both sources to produce more reliable pseudo-labels.

To fuse multi-source bounding boxes, Weighted Box Fusion (WBF)~\cite{solovyev2021weighted} is a straightforward approach that averages overlapping boxes based on confidence scores. However, WBF introduces two critical issues: (1) it requires tuning a global confidence weight, and (2) it merges boxes by class label. This can lead to errors when different sources assign conflicting labels to the same object, which is a frequent scenario due to differing domain perspectives.

To overcome these limitations, we discard class labels during box clustering and instead group overlapping boxes using a large IoU threshold. This allows us to handle label discrepancies more flexibly. For each cluster of overlapping boxes $\{(b_k, p_k)\}_{k=1}^n$, where $b_k \in \mathbb{R}^4$ denotes the box coordinates and $p_k \in [0,1]^C$ the class probability vector, we compute the Shannon entropy $H(p_k)$ of each prediction. We then assign inverse-entropy weights, giving higher influence to more certain predictions. The final fused box and class probability are computed via a weighted average using these normalized weights. The resulting pseudo-label set $\mathcal{P} = \{(\hat{b}_i, \hat{y}_i)\}_{i=1}^{n_p}$ is obtained by selecting the class with the highest fused probability for each box. The fusion procedure is outlined in Algorithm~\ref{alg:depf_simple}, while a more detailed version is provided in the Appendix due to limited space. Through this entropy-aware design, our framework integrates predictions from both the teacher and VFM, yielding pseudo-labels that are more robust to label noise and domain shifts, leading to improved discriminability.

\begin{algorithm}[t]
\caption{DEPF}
\label{alg:depf_simple}
\begin{algorithmic}[1]
\REQUIRE Predictions from two models: $\mathcal{B}_1, \mathcal{B}_2$; IoU threshold $\beta$
\ENSURE Fused predictions $\mathcal{P}$
\STATE Merge all predictions: $\mathcal{B} \leftarrow \mathcal{B}_1 \cup \mathcal{B}_2$
\STATE Group $\mathcal{B}$ into clusters using box IoU $>\beta$
\FORALL{cluster $\mathcal{C}_m = \{(b_k, p_k)\}_{k=1}^n$}
    \STATE Compute entropy-based weights $\tilde{\mathbf{w}}_k \propto 1/H(p_k)$
    \STATE Fuse box: $\hat{b} = \sum_k \tilde{\mathbf{w}}_k b_k$; score: $\hat{p} = \sum_k \tilde{\mathbf{w}}_k p_k$
    \STATE Final label: $\hat{y} = \arg\max_c \hat{p}^{(c)}$
    \STATE Append $(\hat{b}, \hat{y})$ to $\mathcal{P}$
\ENDFOR
\RETURN $\mathcal{P}$
\end{algorithmic}
\end{algorithm}

\subsection{Patch-weighted Global Feature Alignment}
As previously discussed, both discriminability and transferability are crucial for effective adaptation. While DEPF improves discriminability by enhancing pseudo-label quality, we now focus on boosting transferability through feature alignment. To address this, prior SFOD works have aligned features between teacher and student networks using strategies such as graph-based matching~\cite{li2022source} or adversarial learning~\cite{chu2023adversarial}. However, these methods operate solely within the teacher-student paradigm and thus inherit source-induced biases and limited decision boundaries. Vision Foundation Models (VFMs), particularly DINOv2, offer a broader and more transferable feature space due to large-scale pretraining. DT~\cite{lavoie2025large} explores feature alignment between a student backbone and a frozen DINOv2 using patch-level cosine similarity. Yet, it assumes equal contribution from all patches, neglecting the varying semantic importance and domain invariance across regions.

\begin{figure}[!t]
\begin{center}
\includegraphics[width=0.9\linewidth]{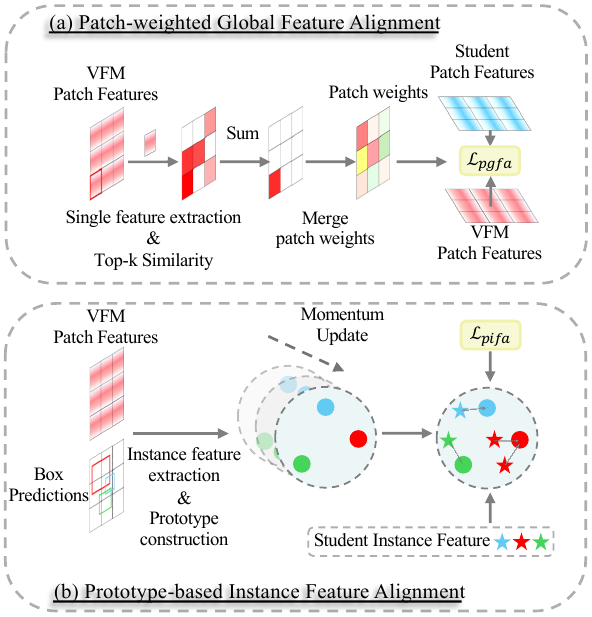}
\caption{PIFA and PGFA. PGFA adopts similarity-based patch weights for fine-grained global feature fusion. PIFA extracts instance feature from VFM to construct a momentum-updated prototype for contrastive alignment with student instance feature.}
\label{fig:alignment}
\end{center}
\end{figure}

\begin{table*}[!t]
\centering
\caption{Results of cross-weather adaptation (Cityscapes to Foggy Cityscapes). }
\small
\resizebox{\linewidth}{!}{
\begin{tabular}{lcc|cccccccc|c}
\toprule
Method & Source-Free & Detector & Truck & Car & Rider & Person & Train & Motor & Bicycle & Bus & mAP \\
\midrule
Source & - & DETR & 15.1 & 46.5 & 39.3 & 38.9 & 4.0 & 21.8 & 36.8 & 34.2 & 29.6 \\
\midrule
CAT~\cite{kennerley2024cat} & \XSolidBrush & \multirow{2}{*}{Faster R-CNN} & 40.8 & 63.7 & 57.1 & 44.6 & 49.7 & 44.9 & 53.0 & 66.0 & 52.5 \\
DT~\cite{lavoie2025large} & \XSolidBrush &  & 47.2 & 65.4 & 60.0 & 48.5 & 52.9 & 46.2 & 56.7 & 66.5 & 55.4 \\
\midrule
MTTrans~\cite{yu2022mttrans} & \XSolidBrush & \multirow{3}{*}{DETR} & 25.8 & 65.2 & 49.9 & 47.7 & 33.9 & 32.6 & 46.5 & 45.9 & 43.4 \\
BiADT~\cite{he2023bidirectional} & \XSolidBrush &  & 31.7 & 69.2 & 58.9 & 52.2 & 45.1 & 42.6 & 51.3 & 55.0 & 50.8 \\
ACCT~\cite{zeng2024enhancing} & \XSolidBrush &  & 31.1 & 69.4 & 58.9 & 53.6 & 33.7 & 42.6 & 54.4 & 53.5 & 49.6 \\
\midrule
IRG-SFDA~\cite{vs2023instance} &  \Checkmark & \multirow{5}{*}{Faster R-CNN} & 24.4 & 51.9 & 45.2 & 37.4 & 25.2 & 31.5 & 41.6 & 39.6 & 37.1 \\
AASFOD~\cite{chu2023adversarial} &  \Checkmark &  & 28.1 & 44.6 & 44.1 & 32.3 & 29.0 & 31.8 & 38.9 & 34.3 & 35.4 \\
BT~\cite{deng2024balanced}& \Checkmark & & 24.3 & 52.7 & 47.1 & 38.4 & 36.3 & 30.2 & 40.1 & 44.6 & 39.2 \\
Simple-SFOD\textsuperscript{\dag}~\cite{hao2024simplifying} &  \Checkmark &  & \underline{30.2} & 54.8 & 44.6 & 36.5 & 31.8 & 29.5 & 41.0 & 45.3 & 39.2 \\
LPLD~\cite{yoon2024enhancing} &  \Checkmark &  & 29.6 & 56.6 & 49.1 & 39.7 & 26.4 & \underline{36.1} & 43.6 & \underline{46.3} & 40.9 \\
\midrule
DRU~\cite{khanh2024dynamic} &  \Checkmark & \multirow{2}{*}{DETR} & 26.2 & \underline{62.5} & \textbf{51.5} & \underline{48.3} & \underline{34.1} & 34.2 & \textbf{48.6} & 43.2 & \underline{43.6} \\
\textbf{Ours} &  \Checkmark &  & \textbf{36.6} & \textbf{62.8} & \underline{51.0} & \textbf{48.8} & \textbf{39.9} & \textbf{36.4} & \underline{48.2} & \textbf{52.7} & \textbf{47.1} \\
\midrule
Oracle & - & DETR & 31.3 & 71.8 & 52.9 & 52.9 & 41.0 & 41.4 & 44.0 & 53.9 & 48.7 \\
\bottomrule
\end{tabular}%
}
\begin{minipage}{\linewidth}
\footnotesize \textsuperscript{\dag} We report results without BN in the backbone following all other approaches for fair comparison.
\end{minipage}
\label{tab:cw}
\end{table*}

To address this, we propose a \textbf{Patch-weighted Global Feature Alignment (PGFA)} module that introduces an adaptive patch-wise weighting scheme. We assign a weight to each patch based on its semantic coherence with other patches in the same image. The calculation proceeds in three steps: First, for each image in a minibatch, we compute the pairwise cosine similarity between all its $L_2$-normalized DINOv2 patch features ($\hat{\mathbf{F}}^{\mathrm{D}}_{b}=\mathbf{F}^{\mathrm{D}}_{b} / \|\mathbf{F}^{\mathrm{D}}_{b}\|_2$, where $\mathrm{D}$ stands for DINOv2 and $b$ stands for batch index). This results in a similarity matrix $\mathbf{S}_b \in \mathbb{R}^{N \times N}$, where $s_{b,i,j} = \langle \hat{\mathbf{F}}^{\mathrm{D}}_{b,i}, \hat{\mathbf{F}}^{\mathrm{D}}_{b,j} \rangle$ and $N\times N$ is the number of patches. Second, we apply a temperature-controlled softmax function row-wise to this matrix to obtain a probability distribution for each patch's similarity to all others. This yields a new matrix $\mathbf{P}_b \in \mathbb{R}^{N \times N}$:
\begin{equation}
    \mathbf{P}_{b,i,j} = \frac{\exp(s_{b,i,j} / \tau)}{\sum_{l=1}^{N} \exp(s_{b,i,l} / \tau)},
\end{equation}
where $\tau=0.07$ is a temperature parameter. Finally, the unnormalized weight $\mathbf{w}_{b,i}$ for patch $i$ is calculated by summing the probabilities of its top-$k$ most similar patches. Let $\mathcal{T}_k(i)$ be the set of indices for the top-$k$ values in the $i$-th row of $\mathbf{P}_b$. The weight $ \mathbf{w}_{b,i} = \sum_{j \in \mathcal{T}_k(i)} p_{b,ij}$ is then normalized across all patches for each image:
\begin{equation}
    \tilde{\mathbf{w}}_{b,i} = \frac{\mathbf{w}_{b,i}}{\sum_{j=1}^{N} \mathbf{w}_{b,j} + \varepsilon}, \quad i = 1,\dots,N,
\end{equation}
where $N$ is the number of patches and $\varepsilon$ ensures numerical stability. This process assigns higher weights to salient, semantically consistent patches, while down-weighting noisy or domain-specific regions. For alignment, following DT, we adopt a weighted cosine loss between $L_2$-normalized patch features from DINOv2 ($\mathbf{F}^{\mathrm{D}}$) and the student model ($\mathbf{F}^{\mathrm{S}}$):
\begin{equation}
\mathcal{L}_{\mathrm{pgfa}} = \frac{1}{B} \sum_{b=1}^{B}
\sum_{i=1}^{N} \tilde{\mathbf{w}}_{b,i}\big(1 - \cos(\mathbf{F}^{\mathrm{D}}_{b,i}, \mathbf{F}^{\mathrm{S}}_{b,i})\big),
\end{equation}
where $B$ is the batch size. Note that in our implementation, we first input student model patch features into a lightweight alignment module \textit{e.g.} MLP. This loss encourages the student to align more strongly with transferable regions in the VFM feature space, improving global feature transferability in a fine-grained manner.

\subsection{Prototype-based Instance Feature Alignment}

While PGFA enhances transferability by aligning global features, accurate object detection further relies on instance-level representations. Given the rich semantic priors in VFMs, we propose leveraging them also as semantic anchors to guide instance recognition and complement global alignment. Specifically, we design \textbf{Prototype-based Instance Feature Alignment (PIFA)}, which constructs class-wise prototypes from VFM features and applies prototype-based contrastive learning to enforce semantic consistency, facilitating more robust instance-level alignment.

\noindent \textbf{Prototype update.}
We first extract a global feature map $\mathbf{F}^{\mathrm{D}} \in \mathbb{R}^{C \times H \times W}$ from the VFM backbone. Given pseudo-labeled boxes ${(b_{k’}, y_{k’})}_{k’=1}^{n’}$, we apply RoIAlign~\cite{he2017mask} to obtain instance features:
\begin{equation}
\mathbf{f}_c^t = \frac{1}{N_c} \sum_{y_{k’}=c} \mathrm{RoIAlign}(\mathbf{F}^{\mathrm{D}}, b_{k’}),
\end{equation}
where $\mathbf{f}_c^t$ is the mean feature for class $c$ in the current batch. 

Inspired by momentum prototype~\cite{li2020mopro}, to maintain stable and evolving class-level semantics, we update the prototype $\mathbf{p}_c^t$ using EMA:
\begin{equation}
\mathbf{p}_c^t = \mu \mathbf{p}_c^{t-1} + (1 - \mu) \mathbf{f}_c^t,
\label{eq:prototype_update}
\end{equation}
where $\mu=0.999$ following previous approaches~\cite{li2020mopro}. This momentum update smooths temporal fluctuations in the class-wise feature representation while gradually incorporating new semantic information.

\noindent \textbf{Contrastive learning.}
To align the feature spaces of the VFM and the student model's instance features, we first pass the patch-level features from the student model through a lightweight alignment module (e.g., an MLP). After projection, both the resulting features $\mathbf{f}_i$ and the prototypes $\mathbf{p}_c$ are $L_2$-normalized to ensure consistent feature magnitudes and facilitate effective similarity computation. Given pseudo label $\hat{y}_i$ obtained in DEPF, the contrastive loss is computed using InfoNCE~\cite{oord2018representation}:
\begin{equation}
\mathcal{L}_{\mathrm{pifa}} = -\frac{1}{M} \sum_{i=1}^{M} \log \frac{\exp\left( \mathbf{f}_i^\top \mathbf{p}_{\hat{y}_i} / \tau \right)}{\sum_{c=1}^{K} \exp\left( \mathbf{f}_i^\top \mathbf{p}_c / \tau \right)},
\end{equation}
where $\tau$ is a temperature hyperparameter and $M$ is the number of non-empty prototypes. This loss encourages each instance-level feature to align closely with its corresponding class prototype while being repelled from other class prototypes. The use of VFM-guided prototypes as anchors ensures that the student model learns instance features that are semantically aligned and more domain-invariant, leading to improved instance-level feature transferability and class-wise discriminability.

\subsection{Overall Objective}
For model optimization, the student is optimized to minimize the weighted summation of the original detection loss $\mathcal{L}_{\mathrm{det}}$ calculated with pseudo labels, and the proposed $\mathcal{L}_{\mathrm{pgfa}}$ and $\mathcal{L}_{\mathrm{pifa}}$ losses:
\begin{equation}
    \mathcal{L}_{\mathrm{tot}} = \mathcal{L}_{\mathrm{det}} + \lambda(\mathcal{L}_{\mathrm{pgfa}} + \mathcal{L}_{\mathrm{pifa}}).
\label{eq:overall_objective}
\end{equation}
The teacher model is then updated through an exponential moving average of student parameters as stated in Eq.~\ref{eq:teacher_update}.

\begin{table}[!t]
\centering

\caption{Results on cross-scene adaptation (Cityscapes to BDD100K). ``FRCNN'' stands for Faster R-CNN}
\resizebox{\linewidth}{!}{
\begin{tabular}{lcc|ccccccc|c}
\toprule
Method & SF & Detector & Truck & Car & Rider & Prsn & Bike & Motor & Bus & mAP \\
\midrule
Source & - & DETR & 17.5 & 57.0 & 29.4 & 43.7 & 15.6 & 17.7 & 17.6 & 28.3 \\
\midrule
AASFOD~\cite{chu2023adversarial} &  \Checkmark & \multirow{2}{*}{FRCNN} & 26.6 & 50.2 & 36.3 & 33.2 & 22.5 & \underline{28.2} & 24.4 & 31.6 \\
BT~\cite{deng2024balanced}& \Checkmark    &       & 24.2  & 50.4  & 34.6  & 32.7  & 28.5  & 24.7  & 24.9  & 31.4  \\
\midrule
DRU~\cite{khanh2024dynamic} &  \Checkmark & \multirow{2}{*}{DETR} & \underline{27.1} & \underline{62.7} & \underline{36.9} & \underline{45.8} & \textbf{32.5} & 22.7 & \underline{28.1} & \underline{36.6} \\
\textbf{Ours} &  \Checkmark &  & \textbf{33.2} & \textbf{72.3} & \textbf{44.2} & \textbf{54.9} & \underline{29.0} & \textbf{32.9} & \textbf{34.8} & \textbf{43.0} \\
\midrule
Oracle & - & DETR & 66.9 & 87.9 & 56.4 & 74.9 & 53.8 & 68.3 & 55.0 & 66.2 \\
\bottomrule
\end{tabular}%
}
\label{tab:c2b}
\end{table}

\begin{table}[!t]
\centering\footnotesize

\caption{Results of synthetic-to-real (Sim10k to Cityscapes, S2R) and cross-scene (KITTI to Cityscapes, K2C) adaptation. We report car AP@50 for each method under each scenario.}
\resizebox{\linewidth}{!}{
\begin{tabular}{lcc|cc}
\toprule
\multirow{2}{*}{Method} & \multirow{2}{*}{Source-Free} & \multirow{2}{*}{Detector} & \multicolumn{2}{c}{mAP} \\
\cmidrule(lr){4-5}
 &  &  & S2R & K2C \\
\midrule
Source & - & DETR & 50.8 & 33.9 \\
\midrule
SFA~\cite{wang2021exploring} & \XSolidBrush & \multirow{2}{*}{DETR} & 52.6 & 46.7 \\
DA-DETR~\cite{zhang2023detr} & \XSolidBrush &  & 54.7 & 48.9 \\
\midrule
IRG-SFDA~\cite{vs2023instance} & \Checkmark & \multirow{4}{*}{Faster R-CNN} & 46.9 & 45.2 \\
AASFOD~\cite{chu2023adversarial} & \Checkmark &  & 44.9 & 44.0 \\
SF-UT~\cite{hao2024simplifying} & \Checkmark &  & 55.4 & 46.2 \\
LPLD~\cite{yoon2024enhancing} & \Checkmark &  & 49.4 & \underline{51.3} \\
\midrule
DRU~\cite{khanh2024dynamic} & \Checkmark & \multirow{2}{*}{DETR} & \underline{58.7} & 45.1 \\
\textbf{Ours} & \Checkmark &  & \textbf{67.4} & \textbf{54.7} \\
\midrule
Oracle & - & DETR & 75.9 & 75.9 \\
\bottomrule
\end{tabular}%
}
\label{tab:s2r-k2c}
\end{table}

\section{Experiments}

\subsection{Datasets and Settings}
We evaluate our method on six widely-used object detection datasets: Cityscapes, Foggy Cityscapes, Sim10k, KITTI, BDD100K, and ACDC. We conclude the settings into three domain adaptation scenarios: cross-weather, synthetic-to-real, and cross-scene.

\noindent \textbf{Cross-weather Adaptation.} Cityscapes~\cite{cordts2016cityscapes} contains 2,975 training and 500 validation images of urban scenes. Its synthetic variant, Foggy Cityscapes~\cite{sakaridis2018semantic}, overlays fog at varying densities; we follow prior work and adopt the 0.02 fog level. Additionally, ACDC~\cite{sakaridis2021acdc} includes four challenging conditions (snow, rain, night, and fog) to further evaluate robustness under diverse weather.

\noindent \textbf{Synthetic-to-Real Adaptation.} Sim10k~\cite{johnson2016driving}, generated from the GTA V engine, provides 9,000 training and 1,000 validation images and serves as the synthetic source. We use Cityscapes as the real-world target to assess synthetic to real generalization.

\noindent \textbf{Cross-scene Adaptation.} We consider two cross-scene transfers: First, following prior work, we transfer from Cityscapes to the daytime split of BDD100K~\cite{yu2018bdd100k} (36,728 training and 5,258 validation images). Second, we transfer from KITTI~\cite{geiger2012we} (7,481 labeled images) to Cityscapes. These two settings test the model’s ability to adapt across cities, camera perspectives, and environmental variations.

\subsection{Implementation Details}
We set $\mu=0.9$ in Eq.~\ref{eq:prototype_update} and $\lambda=1$ in Eq.~\ref{eq:overall_objective}.  $\alpha$ in Eq.~\ref{eq:teacher_update} is set to $0.999$ with EMA applied every 5 iterations. We train for 30 epochs with a learning rate of $5\times10^{-5}$ and batch size of 8. See Appendix for more details.

\subsection{Comparisons with State-of-the-art}

We evaluate our method under various SFOD scenarios using Deformable DETR as the base detector. In the cross-weather setting, our approach achieves 47.1\% mAP, surpassing DRU~\cite{khanh2024dynamic} (43.6\%) and LPLD~\cite{yoon2024enhancing} (40.9\%), with especially large gains in challenging classes such as \texttt{Truck} (+10.4\%) and \texttt{Bus} (+9.5\%). On the Cityscapes-to-ACDC benchmark, our method consistently outperforms all baselines across all weather conditions. For cross-scene adaptation, we achieve a 6.4\% mAP gain over DRU on Cityscapes-to-BDD100K, effectively leveraging the large-scale unlabeled target data. Our performance is also competitive with the non-source-free DT~\cite{lavoie2025large}, trailing by only 4.8\% despite not using source data. On Cityscapes-to-KITTI, we outperform all prior SFOD and DAOD methods by a notable margin. In the Sim10k-to-Cityscapes (synthetic-to-real) setting, our method improves car AP by 8.5\% over DRU and 12.7\% over DA-DETR, demonstrating strong domain adaptation capability across diverse settings.

\begin{table}[!t]
\centering\footnotesize
\caption{Ablation study on component analysis. Mean Teacher baseline is reproduced with our hyperparameter setting for fair comparison.}
\small
\resizebox{\linewidth}{!}{
\begin{tabular}{cccc|c}
\toprule
Mean Teacher~\cite{tarvainen2017mean} & PGFA & PIFA & DEPF & mAP / Gain\\
\midrule
\Checkmark &        &         &       & 42.3 / - \\
\Checkmark & \Checkmark &         &       & 43.4 / +1.1 \\
\Checkmark &        & \Checkmark &       & 43.9 / +1.6 \\
\Checkmark & \Checkmark & \Checkmark &       & 45.0 / +2.7 \\
\Checkmark &        &         & \Checkmark & 45.9 / +3.6 \\
\Checkmark & \Checkmark &         & \Checkmark & 46.3 / +4.0 \\
\Checkmark &        & \Checkmark & \Checkmark & 46.5 / +4.2 \\
\midrule
\Checkmark & \Checkmark & \Checkmark & \Checkmark(w/o $\tilde{\mathbf{w}}_k$) & 46.8 / +4.5 \\
\Checkmark & \Checkmark(w/o $\tilde{\mathbf{w}}_{b}$) & \Checkmark & \Checkmark & 46.5 / +4.2 \\
\midrule
\Checkmark & \Checkmark & \Checkmark & \Checkmark & \textbf{47.1} / +\textbf{4.8} \\
\bottomrule
\end{tabular}
}
\label{tab:abl_components_mAP}
\end{table}

\begin{table}[!t]
\centering
\caption{Results of cross-weather adaptation (Cityscapes-to-ACDC) on Snow, Rain, Night, Fog. ``FRCNN'' stands for Faster R-CNN. ``SF'' stands for Source-free. We report mAP@50 for each method under each scenario.}
\label{tab:city_to_acdc}
\resizebox{\linewidth}{!}{
\begin{tabular}{l|c|l|cccc}
\toprule
Method & SF & Detector & Snow & Rain & Night & Fog \\
\midrule
AT~\cite{li2022at}           & \XSolidBrush & \multirow{2}{*}{FRCNN} & 55.2 & 37.7 & 29.5 & 62.2 \\
DT~\cite{lavoie2025large}     & \XSolidBrush &  & 56.8 & 39.0 & 36.4 & 68.6 \\
\midrule
DRU~\cite{khanh2024dynamic}     & \Checkmark & \multirow{2}{*}{DETR} & \underline{37.9} & \underline{26.3} & \underline{16.5} & \underline{45.4} \\
\textbf{Ours}           & \Checkmark &  & \textbf{47.9} & \textbf{32.1} & \textbf{23.0} & \textbf{54.0} \\
\bottomrule
\end{tabular}%
}
\end{table}

\begin{figure*}[t] 
  \centering
   \begin{overpic}[width=1.0\textwidth]{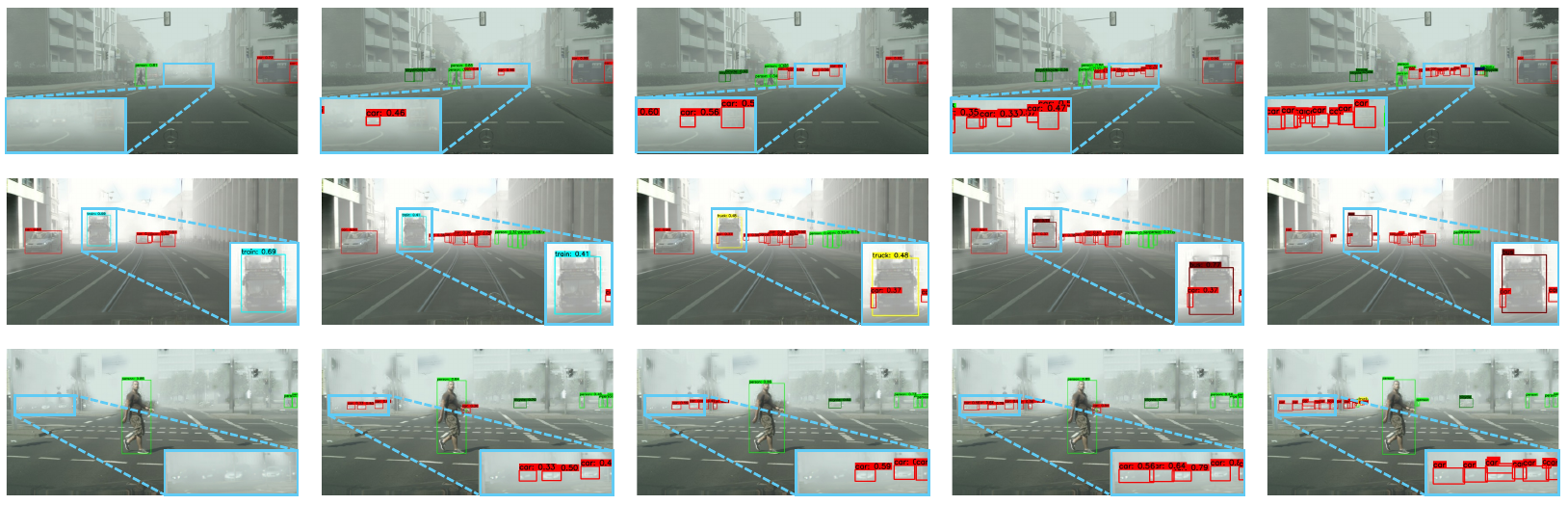} 
   \put(4,-1){(a) Source only}
   \put(23,-1){(b) Mean Teacher}
   \put(46,-1){(c) DRU}
   \put(67,-1){(d) Ours}
   \put(83,-1){(e) Ground Truth}
   \end{overpic}
  \caption{Detection visualization on the Cityscapes-to-Foggy Cityscapes adaptation scenario among (a) Source only, (b) Mean Teacher, (c) DRU, (d) Ours, (e) Ground Truth. We zoom in the discriminative regions for each set of predictions.}
  \label{vis:detection}
\end{figure*}

\begin{figure}[!t] \centering
   
    \includegraphics[width=0.15\textwidth]{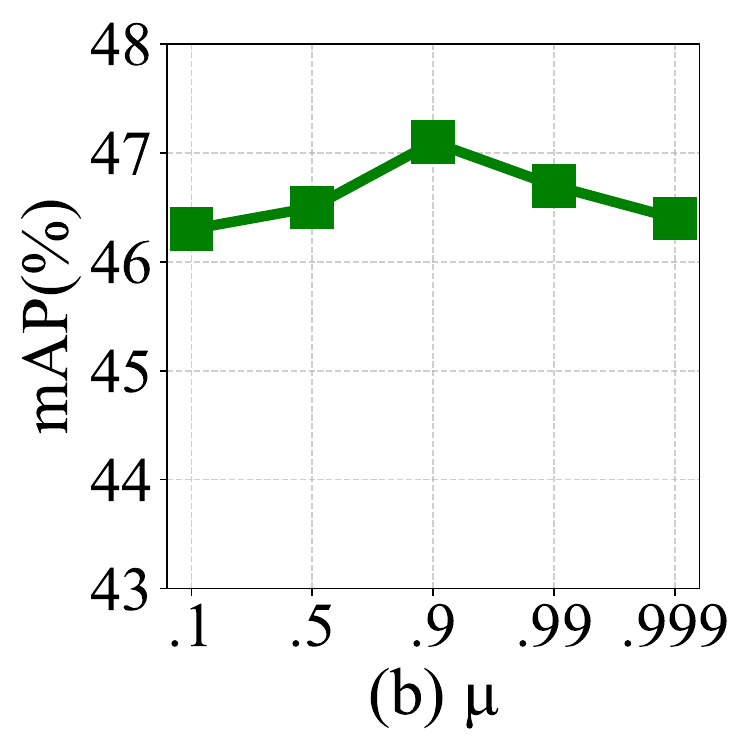}
    \includegraphics[width=0.15\textwidth]{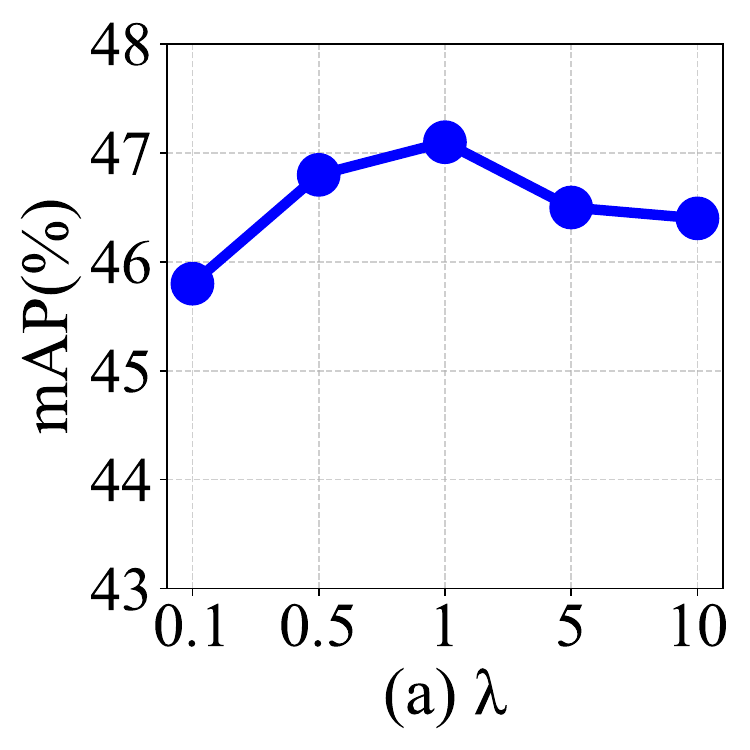}
    \includegraphics[width=0.15\textwidth]{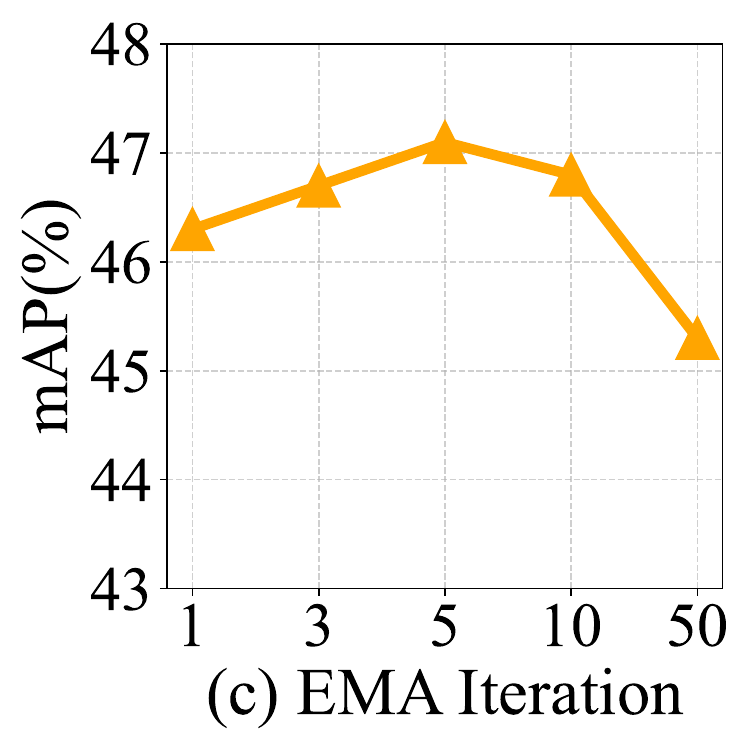}
    \caption{Hyperparameter sensitivity. $\mu$ is the momentum for prototype update, $\lambda$ is the loss balancing factor, EMA iteration is the number of iterations for every EMA update.} \label{fig:ablation_hyperparam}

\end{figure}

\subsection{Ablation Study}
\noindent We conduct ablation experiments are conducted on the Cityscapes-to-Foggy Cityscapes setting, consistent with the main experiments.

\noindent \textbf{Component Effectiveness.} We evaluate each proposed module, including PGFA, PIFA, and DEPF. As shown in Tab.~\ref{tab:abl_components_mAP}, each component brings noticeable gains over the Mean Teacher (MT) baseline, and their combination leads to 47.1\% mAP, a 4.8\% improvement over MT. This confirms their complementary roles in improving transferability and discriminability. Additionally, we assess the effect of the proposed weights: patch-wise distillation weight $\tilde{w}_{b}$ in PGFA and label fusion weight $\tilde{w}_k$ in DEPF. Including these weights yields further gains of 0.6\% and 0.3\% mAP, respectively, highlighting the importance of emphasizing informative regions and predictions.

\noindent \textbf{Hyperparameter Sensitivity.} We evaluate three key hyperparameters: prototype momentum $\mu$ (Eq.~\ref{eq:prototype_update}), loss balancing factor $\lambda$ (Eq.~\ref{eq:overall_objective}), and the EMA update interval, \textit{i.e.}, the number of iterations between teacher updates. As shown in Fig.~\ref{fig:ablation_hyperparam}, our method remains robust across a broad range of settings. However, updating prototypes too frequently or too rarely degrades performance due to pseudo-label noise. An EMA interval of 5 yields the best results, balancing stability and timely adaptation, consistent with prior studies~\cite{chu2023adversarial,zhao2024multi}.

\subsection{Visualization}
To qualitatively assess the effectiveness of our approach, we conduct visualization experiments on detection results. As shown in Fig.~\ref{vis:detection}, our method demonstrates clear advantages over the baselines. It successfully detects challenging objects and produces correct classifications for visually ambiguous instances (line 2, where \textit{bus} is misclassified as \textit{train} by MT and as \textit{truck} by DRU), highlighting the benefits of our method and the improved optimization enabled by reliable pseudo-labels. We also present pseudo-label visualizations in the Appendix.

\subsection{Additional Experiments}
Additional experiments, such as detector variants, zero-shot Grounding DINO performance, efficiency analysis, etc., are provided in the Appendix due to space limitations.

\section{Conclusion}
This paper explores an under-investigated direction of enhancing Source-Free Object Detection (SFOD) using Vision Foundation Models (VFMs) to improve both discriminability and transferability in self-training pipelines. We introduce a simple yet effective framework consisting of three modules: PGFA, PIFA, and DEPF, each focusing on global feature alignment, instance-level feature alignment, and pseudo-label refinement, respectively. By effectively integrating generic knowledge from VFMs, our framework achieves consistent improvements in SFOD performance, as validated by extensive experiments. These results indicate that integrating VFMs into more scalable and robust DAOD and SFOD frameworks is a promising direction for future research. In future work, we aim to explore more about richer vision-language multimodal feature spaces to enhance alignment effectiveness and to incorporate VFMs into more challenging cross-domain object detection scenarios.

\setcounter{secnumdepth}{0}
\section{Acknowledgments}
This work was supported in part by the National Key R\&D Program of China (Grant No.2023YFF0725001), in part by the National Natural Science Foundation of China (Grant No.92370204), in part by the Guangdong Basic and Applied Basic Research Foundation (Grant No.2023B1515120057), in part by the Key-Area Special Project of Guangdong Provincial Ordinary Universities(2024ZDZX1007), in part by the Education Bureau of Guangzhou.

\bibliography{main}

\clearpage
\appendix
\setcounter{secnumdepth}{2}
\maketitle


\section{Detailed Algorithm for DEPF}
Here we provide the detailed algorithm for DEPF (Dual-source Enhanced Pseudo-label Fusion), as shown in Algorithm~\ref{alg:fusion}.

\begin{table*}[!t]
\centering

\caption{Results on cross-scene adaptation (Cityscapes to BDD100K). }
\resizebox{\linewidth}{!}{
\begin{tabular}{lcc|ccccccc|c}
\toprule
Method & Source-Free & Detector & Truck & Car & Rider & Person & Bicycle & Motor & Bus & mAP \\
\midrule
Source & - & DETR & 17.5 & 57.0 & 29.4 & 43.7 & 15.6 & 17.7 & 17.6 & 28.3 \\
\midrule
CAT~\cite{kennerley2024cat} & \XSolidBrush & \multirow{2}{*}{Faster R-CNN} & 31.4 & 61.2 & 41.5 & 44.6 & 31.7 & 24.4 & 34.6 & 38.5 \\
DT~\cite{lavoie2025large} & \XSolidBrush &  & 44.3 & 66.6 & 47.0 & 51.6 & 40.8 & 38.3 & 45.9 & 47.8 \\
\midrule
SFA~\cite{wang2021exploring} & \XSolidBrush & \multirow{5}{*}{DETR} & 19.1 & 57.5 & 27.6 & 40.2 & 19.2 & 15.4 & 23.4 & 28.9 \\
MTTrans~\cite{yu2022mttrans} & \XSolidBrush &  & 25.1 & 61.5 & 30.1 & 44.1 & 23.0 & 17.7 & 26.9 & 32.6 \\
MTM~\cite{weng2024mean} & \XSolidBrush &  & 23.0 & 68.8 & 35.1 & 53.7 & 28.0 & 23.8 & 28.8 & 37.3 \\
BiADT~\cite{he2023bidirectional} & \XSolidBrush &  & 17.4 & 60.9 & 34.0 & 42.1 & 25.7 & 18.2 & 19.5 & 31.1 \\
ACCT~\cite{zeng2024enhancing} & \XSolidBrush &  & 26.0 & 61.8 & 41.4 & 51.8 & 36.9 & 31.7 & 23.4 & 39.0 \\
\midrule
SED~\cite{li2021free} &  \Checkmark & \multirow{3}{*}{Faster R-CNN} & 20.6 & 50.4 & 32.6 & 32.4 & 25.0 & 18.9 & 23.4 & 29.0 \\
AASFOD~\cite{chu2023adversarial} &  \Checkmark &  & 26.6 & 50.2 & 36.3 & 33.2 & 22.5 & 28.2 & 24.4 & 31.6 \\
BT~\cite{deng2024balanced} &  \Checkmark &  & 24.2 & 50.4 & 34.6 & 32.7 & 28.5 & \underline{24.7} & 24.9 & 31.4 \\
\midrule
DRU~\cite{khanh2024dynamic} &  \Checkmark & \multirow{2}{*}{DETR} & \underline{27.1} & \underline{62.7} & \underline{36.9} & \underline{45.8} & \textbf{32.5} & 22.7 & \underline{28.1} & \underline{36.6} \\
\textbf{Ours} &  \Checkmark &  & \textbf{33.2} & \textbf{72.3} & \textbf{44.2} & \textbf{54.9} & \underline{29.0} & \textbf{32.9} & \textbf{34.8} & \textbf{43.0} \\
\midrule
Oracle & - & DETR & 66.9 & 87.9 & 56.4 & 74.9 & 53.8 & 68.3 & 55.0 & 66.2 \\
\bottomrule
\end{tabular}%
}
\label{tab:c2b_full}
\end{table*}

\begin{table*}[!t]
\centering
\caption{Comparison of source-free adaptation across different detectors on Foggy Cityscapes. For each detector group, the best per-category AP is highlighted in bold.}
\resizebox{\linewidth}{!}{
\begin{tabular}{lcccccccccc}
\toprule
Method & Detector & Truck & Car & Rider & Person & Train & Motor & Bicycle & Bus & mAP \\
\midrule
Source\textsuperscript{\dag} & \multirow{3}{*}{Faster R-CNN~\cite{ren2016faster}} & 19.8 & 36.1 & 38.6 & 31.2 & 9.1 & 21.8 & 30.4 & 23.5 & 26.3 \\
IRG\textsuperscript{\dag}~\cite{vs2023instance} & & 23.8 & 51.4 & 46.5 & 37.5 & 30.3 & 28.7 & 39.5 & 39.5 & 37.1 \\
IRG+Ours & & \textbf{29.0} & \textbf{51.7} & \textbf{48.0} & \textbf{37.6} & \textbf{35.4} & \textbf{32.8} & \textbf{42.7} & \textbf{46.8} & \textbf{40.5} \\
\midrule
Source & \multirow{3}{*}{RT DETR~\cite{zhao2024detrs}} & 29.9 & 54.0 & 46.8 & 38.7 & 21.9 & 29.4 & 39.3 & 43.7 & 38.0 \\
MT~\cite{tarvainen2017mean} & & 40.2 & 71.2 & 52.0 & 48.8 & 47.7 & 40.8 & 45.3 & 61.6 & 51.0 \\
MT+Ours & & \textbf{43.8} & \textbf{73.9} & \textbf{54.3} & \textbf{51.6} & \textbf{53.7} & \textbf{42.2} & \textbf{48.6} & \textbf{66.3} & \textbf{54.3} \\
\midrule
Source\textsuperscript{\dag} & \multirow{3}{*}{YOLOv5~\cite{glenn_jocher_2020_4154370}} & 25.3 & 59.9 & 49.1 & 48.1 & 18.8 & 33.2 & 41.1 & 32.8 & 38.5 \\
SF-YOLO\textsuperscript{\dag}~\cite{varailhon2024source} & & 30.0 & \textbf{62.4} & \textbf{53.1} & \textbf{47.9} & 34.4 & 36.9 & 43.4 & 41.8 & 43.7 \\
SF-YOLO+Ours & & \textbf{31.1} & 62.0 & 53.0 & 47.7 & \textbf{45.9} & \textbf{39.9} & \textbf{44.9} & \textbf{44.5} & \textbf{46.1} \\
\bottomrule
\end{tabular}
}
\begin{minipage}{\linewidth}
\footnotesize \textsuperscript{\dag} We reproduce the results based on the corresponding publicly available codebases.
\end{minipage}
\label{tab:sfod_detector_comparison}
\end{table*}

\begin{table*}[!t]
\centering
\caption{Per-category AP results across different methods and backbones on Cityscapes-to-Foggy Cityscapes scenario.}
\resizebox{\linewidth}{!}{
\begin{tabular}{lc|cccccccc|c}
\toprule
Method & Backbone & Truck & Car & Rider & Person & Train & Motor & Bicycle & Bus & mAP \\
\midrule
Grounding DINO~\cite{liu2024grounding} & Swin-B & 32.8 & 49.7 & 16.7 & 33.4 & 19.4 & 33.3 & 42.2 & 46.5 & 34.3 \\
\midrule
Source & \multirow{3}{*}{ResNet50} & 15.1 & 46.5 & 39.3 & 38.9 & 4.0 & 21.8 & 36.8 & 34.2 & 29.6 \\
DRU~\cite{khanh2024dynamic}    &          & 26.2 & 62.5 & \textbf{51.5} & 48.3 & 34.1 & 34.2 & \textbf{48.6} & 43.2 & 43.6 \\
Ours   &          & \textbf{36.6} & \textbf{62.8} & 51.0 & \textbf{48.8} & \textbf{39.9} & \textbf{36.4} & 48.2 & \textbf{52.7} & \textbf{47.1} \\
\midrule
Source & \multirow{3}{*}{ViT-B}    & 17.8 & 46.8 & 41.3 & 38.2 & 3.7 & 24.2 & 37.2 & 30.1 & 29.9 \\
DRU~\cite{khanh2024dynamic}    &          & 22.5 & 62.6 & 45.9 & 44.8 & 20.9 & 25.5 & 41.1 & 39.2 & 37.8 \\
Ours   &          & \textbf{33.7} & \textbf{63.2} & \textbf{52.5} & \textbf{49.9} & \textbf{34.4} & \textbf{37.0} & \textbf{47.8} & \textbf{47.4} & \textbf{45.7} \\
\midrule
Source & \multirow{3}{*}{Swin-T}   & 30.0 & 55.5 & 48.5 & 45.0 & 15.8 & 35.3 & 42.6 & 41.3 & 39.2 \\
DRU~\cite{khanh2024dynamic}    &          & 35.0 & \textbf{66.4} & 49.7 & 49.8 & 30.4 & 38.1 & 47.6 & 46.9 & 45.5 \\
Ours   &          & \textbf{40.0} & \textbf{66.4} & \textbf{53.1} & \textbf{51.4} & \textbf{34.4} & \textbf{41.7} & \textbf{50.8} & \textbf{46.8} & \textbf{48.1} \\
\midrule
Source & \multirow{3}{*}{Swin-S}   & 29.9 & 56.2 & 50.5 & 45.6 & \textbf{35.9} & 36.6 & 42.7 & 46.6 & 43.0 \\
DRU~\cite{khanh2024dynamic}    &          & 37.8 & 65.8 & 53.8 & 50.8 & 27.6 & 41.8 & 49.2 & 50.7 & 47.1 \\
Ours   &          & \textbf{40.9} & \textbf{66.2} & \textbf{55.3} & \textbf{52.2} & 31.3 & \textbf{44.1} & \textbf{50.8} & \textbf{54.4} & \textbf{49.4} \\
\midrule
Source & \multirow{3}{*}{Swin-B}   & 37.7 & 62.5 & 51.3 & 49.3 & 39.2 & 38.6 & 45.2 & 46.2 & 46.2 \\
DRU~\cite{khanh2024dynamic}    &          & 39.4 & \textbf{69.2} & 52.5 & 51.8 & 42.1 & 44.7 & 50.8 & 56.4 & 50.9 \\
Ours   &          & \textbf{43.4} & 69.1 & \textbf{55.0} & \textbf{53.8} & \textbf{44.8} & \textbf{44.6} & \textbf{53.1} & \textbf{59.3} & \textbf{52.9} \\
\midrule
Source & \multirow{3}{*}{Swin-L}   & 39.4 & 63.5 & 49.5 & 49.2 & \textbf{48.9} & 41.9 & 45.6 & 52.7 & 48.9 \\
DRU~\cite{khanh2024dynamic}    &          & \textbf{45.0} & 68.8 & 52.9 & 53.1 & 47.8 & 45.2 & 50.5 & 56.6 & 52.5 \\
Ours   &          & 44.9 & \textbf{69.2} & \textbf{57.1} & \textbf{55.0} & 44.1 & \textbf{47.6} & \textbf{53.5} & \textbf{57.9} & \textbf{53.7} \\
\bottomrule
\end{tabular}%
}
\label{tab:backbone_full}
\end{table*}

\begin{table*}[!t]
\centering
\caption{Detailed results of cross-weather adaptation on Cityscapes-to-ACDC benchmark.}
\resizebox{\linewidth}{!}{
\begin{tabular}{lcc|cccccccc|c}
\toprule
Method & Backbone & Scenario & Truck & Car & Rider & Person & Train & Motor & Bicycle & Bus & mAP \\
\midrule
\multirow{4}{*}{DRU~\cite{khanh2024dynamic}} 
& \multirow{4}{*}{ResNet50} & Snow  & 44.2 & 62.2 & 11.7 & 37.7 & 64.5 & 49.1 & 26.1 & 8.0  & 37.9 \\
& & Rain  & 17.8 & 59.9 & 11.8 & 35.8 & 27.4 & 36.7 & 0.0  & 21.0 & 26.3 \\
& & Night & 18.6 & 40.7 & 5.3  & 24.7 & 11.4 & 11.9 & 3.2  & -\textsuperscript{\dag} & 16.5 \\
& & Fog   & 23.9 & 64.6 & 35.5 & 50.0 & 66.3 & 44.1 & 44.0 & 34.5 & 45.4 \\
\midrule
\multirow{4}{*}{Grounding DINO~\cite{liu2024grounding}} 
& \multirow{4}{*}{Swin-B} & Snow  & 46.6 & 78.0 & 0.0 & 66.5 & 68.2 & 48.7 & 64.5 & 36.4  & 51.1 \\
& & Rain  & 53.0 & 79.7 & 5.4 & 41.9 & 22.8 & 41.4 & 16.8  & 37.5 & 37.3 \\
& & Night & 21.4 & 61.5 & 3.1 & 42.3 & 46.4 & 10.3 & 10.3 & -\textsuperscript{\dag} & 27.9 \\
& & Fog   & 38.3 & 84.6 & 18.8 & 61.9 & 66.3 & 46.2 & 61.4 & 71.7 & 56.2 \\
\midrule
\multirow{4}{*}{Ours} 
& \multirow{4}{*}{ResNet50} & Snow  & 48.8 & 72.2 & 51.3 & 42.6 & 69.7 & 55.0 & 42.1 & 1.5  & 47.9 \\
& & Rain  & 27.8 & 71.4 & 31.2 & 39.4 & 21.0 & 23.3 & 9.5  & 32.8 & 32.1 \\
& & Night & 14.3 & 55.0 & 18.3 & 33.6 & 17.3 & 11.5 & 11.0 & -\textsuperscript{\dag} & 23.0 \\
& & Fog   & 35.6 & 80.3 & 52.0 & 55.4 & 66.3 & 36.6 & 40.0 & 65.6 & 50.6 \\
\midrule
\multirow{4}{*}{Ours} 
& \multirow{4}{*}{Swin-B} & Snow  & 58.0 & 73.6 & 51.1 & 51.7 & 46.9 & 69.1 & 57.7 & 8.9  & 52.1 \\
& & Rain  & 47.0 & 72.4 & 21.6 & 49.0 & 40.1 & 43.0 & 7.7  & 40.5 & 40.1 \\
& & Night & 33.2 & 55.8 & 12.8 & 33.2 & 36.4 & 22.5 & 17.5 & -\textsuperscript{\dag} & 30.2 \\
& & Fog   & 39.1 & 81.5 & 50.4 & 65.3 & 66.3 & 57.6 & 63.9 & 81.0 & 63.1 \\
\bottomrule
\end{tabular}
}

\begin{minipage}{\linewidth}
\footnotesize \textsuperscript{\dag} No \texttt{bus} instance is available in the ACDC-night validation split.
\end{minipage}
\label{tab:acdc_full}
\end{table*}

\section{Implementation Details}

In this section, we provide the complete implementation details of our method.

For the source model pretraining stage, we train for 40 epochs with a learning rate of $2 \times 10^{-4}$, followed by another 10 epochs with a reduced learning rate of $2 \times 10^{-5}$. During the target adaptation stage, we set the total batch size to 8 and use a learning rate of $5 \times 10^{-5}$. We set the EMA decay factor $\alpha$ to 0.999 and perform the teacher model update every 5 iterations, ensuring stable updates as suggested by prior works~\cite{chu2023adversarial,zhao2024multi}. For other hyperparameters, we adopt the default settings discussed in the main paper. Specifically, we set $\mu = 0.9$ as the momentum coefficient in prototype updates, and $\lambda = 1$ as the balancing weight in the overall objective. For the DEPF module, we set the IoU threshold $\beta = 0.7$ to cluster and fuse overlapping bounding boxes. In PGFA, we select the top 50 patches with the highest similarity scores to construct patch-wise weights. Both student and VFM features are extracted using $16 \times 16$ patch tokens. For RoIAlign~\cite{he2017mask}, we fix the output resolution to $7 \times 7$, consistent with most standard implementations. We use a temperature parameter of $\tau = 0.07$ for the contrastive loss in PIFA, following prior practices~\cite{wang2021understanding}. The lightweight projection heads used in both PGFA and PIFA are three-layer MLPs with a hidden dimension of 512. The input and output dimensions are determined by the channel dimensions of the student detector, student backbone, and visual foundation models.

Our framework is implemented using PyTorch~\cite{paszke2019pytorch}. All experiments are conducted on a machine with 4 NVIDIA RTX A6000 GPUs running Ubuntu 22.04 as the operating system.

\begin{algorithm}[t]
\caption{DEPF}
\label{alg:fusion}
\begin{algorithmic}[1]
\REQUIRE Two sets of predictions $\mathcal{B}_1 = \{(b_i, p_i)\}$, $\mathcal{B}_2 = \{(b_j, p_j)\}$; IoU threshold $\beta$; small constant $\varepsilon$
\ENSURE Fused pseudo labels $\{(\hat{b}_{k'}, \hat{y}_{k'})\}$

\STATE Merge predictions: $\mathcal{B} \leftarrow \mathcal{B}_1 \cup \mathcal{B}_2$
\STATE Initialize empty cluster list: $\mathcal{C} \leftarrow []$
\FORALL{$(b_k, p_k) \in \mathcal{B}$}
    \STATE Find cluster $\mathcal{C}_m$ such that $\exists (b_j, \cdot) \in \mathcal{C}_m$ with $\mathrm{IoU}(b_k, b_j) > \beta$
    \IF{such cluster exists}
        \STATE Add $(b_k, p_k)$ to cluster $\mathcal{C}_m$
    \ELSE
        \STATE Create new cluster $\mathcal{C}_{\text{new}} \leftarrow \{(b_k, p_k)\}$ and add to $\mathcal{C}$
    \ENDIF
\ENDFOR

\STATE Initialize fused label list: $\mathcal{P} \leftarrow []$
\FORALL{cluster $\mathcal{C}_m = \{(b_k, p_k)\}_{k=1}^{n}$}
    \FOR{$k = 1$ to $n$}
        \STATE $H(p_k) \leftarrow -\sum_{c} p_k^{(c)} \log(p_k^{(c)} + \varepsilon)$
        \STATE $\mathbf{w}_k \leftarrow \frac{1}{H(p_k) + \varepsilon}$
    \ENDFOR
    \STATE Normalize weights: $\tilde{\mathbf{w}}_k \leftarrow \mathrm{Norm}(\mathbf{w}_k)$
    \STATE Fuse box: $\hat{b} \leftarrow \sum_{k=1}^{n} \tilde{\mathbf{w}}_k b_k$
    \STATE Fuse score: $\hat{p} \leftarrow \sum_{k=1}^{n} \tilde{\mathbf{w}}_k p_k$
    \STATE Final label: $\hat{y} \leftarrow \arg\max_c \hat{p}^{(c)}$
    \STATE Append $(\hat{b}, \hat{y})$ to $\mathcal{P}$
\ENDFOR

\RETURN $\mathcal{P}$
\end{algorithmic}
\end{algorithm}

\section{Full Experimental Results}
In the main paper, we only report partial results for two adaptation scenarios including (1) Cityscapes-to-BDD100K and (2) Cityscapes-to-ACDC, due to space limitations. Here, we provide the complete results for both settings.

\subsection{Cityscapes-to-BDD100K}
For the Cityscapes-to-BDD100K scenario, we include a more comprehensive comparison with additional baseline methods and experimental settings, as shown in Tab.~\ref{tab:c2b_full}. The results demonstrate that our approach consistently surpasses most existing DAOD and source-free object detection (SFOD) methods by a substantial margin—for instance, achieving 6.4\% higher mAP than DRU~\cite{khanh2024dynamic} and 4.0\% higher than ACCT~\cite{zeng2024enhancing}. While our method performs slightly worse than DT~\cite{lavoie2025large}, it is important to note that DT assumes access to both source images and labels, making its setting less constrained than ours. Since DT also incorporates Vision Foundation Models (VFMs), its strong performance further highlights the potential of VFMs for enhancing cross-domain generalization.

\subsection{Cityscapes-to-ACDC}
For the Cityscapes-to-ACDC setting, we report detailed class-wise AP and overall mAP for both DRU~\cite{khanh2024dynamic} and our method, based on our own reproduction and implementation (Tab.~\ref{tab:acdc_full}). The results clearly show that our method consistently outperforms DRU across most categories, further validating the robustness and effectiveness of our approach.

\section{Additional Experiments}
\subsection{Effectiveness across Backbone Variants}
We further evaluate the method across different detector backbones beyond ResNet-50, including Swin Transformer~\cite{liu2021swin} with sizes T(tiny), S(small), B(base), and L(large) and Vision Transformer ViT-B~\cite{dosovitskiy2020image}. We give the full results, including per-class AP and mAP. Results in Tab.~\ref{tab:backbone_full} show consistent gains across all backbones, verifying the robustness of our method across different backbones. These gains demonstrate that our approach not only scales well with model capacity but also complements different feature extraction paradigms, whether convolution-based (ResNet), hierarchical transformer-based (Swin), or pure transformer-based (ViT). Importantly, even with lightweight backbones such as Swin-T, our method maintains strong performance, making it suitable for resource-constrained deployment scenarios as well.

\subsection{Effectiveness across Detector Variants}
Although our main experiments are conducted using Deformable DETR~\cite{zhu2020deformable}, our method is designed to be detector-agnostic, meaning it can be readily adapted to other detection architectures. To demonstrate this generality, we implement our method on three representative detector variants: Faster R-CNN~\cite{ren2016faster}, RT-DETR~\cite{zhao2024detrs}, and YOLOv5~\cite{glenn_jocher_2020_4154370}.

For ease of implementation and fair comparison, we build on publicly available open-source baselines: we use the IRG~\cite{vs2023instance} framework for Faster R-CNN and SF-YOLO~\cite{varailhon2024source} for YOLOv5 (specifically, the YOLOv5l variant). For RT-DETR, we implement a simple Mean Teacher (MT)~\cite{tarvainen2017mean} framework as baseline. To avoid conflicts between components, we make the following adjustments:
\begin{itemize}
    \item For IRG+ours, we disable the instance relation graph and the graph-based contrastive loss, as they may interfere with our PIFA module’s instance-level contrastive learning.
    \item For RT-DETR, which relies on ground-truth labels for query selection during training, we disable this mechanism when generating pseudo-labels from the teacher model. Instead, pseudo-labels are used to guide query selection in the student model during training.
\end{itemize}

As shown in Tab.~\ref{tab:sfod_detector_comparison}, our method achieves consistent and significant performance gains across all detector variants: +3.4\% mAP on Faster R-CNN, +3.3\% on RT-DETR, and +2.4\% on YOLOv5l. These results confirm the flexibility and effectiveness of our approach across a range of detection paradigms, underscoring its potential for real-world deployment.

\begin{figure}[!t]
\begin{center}
\includegraphics[width=\linewidth]{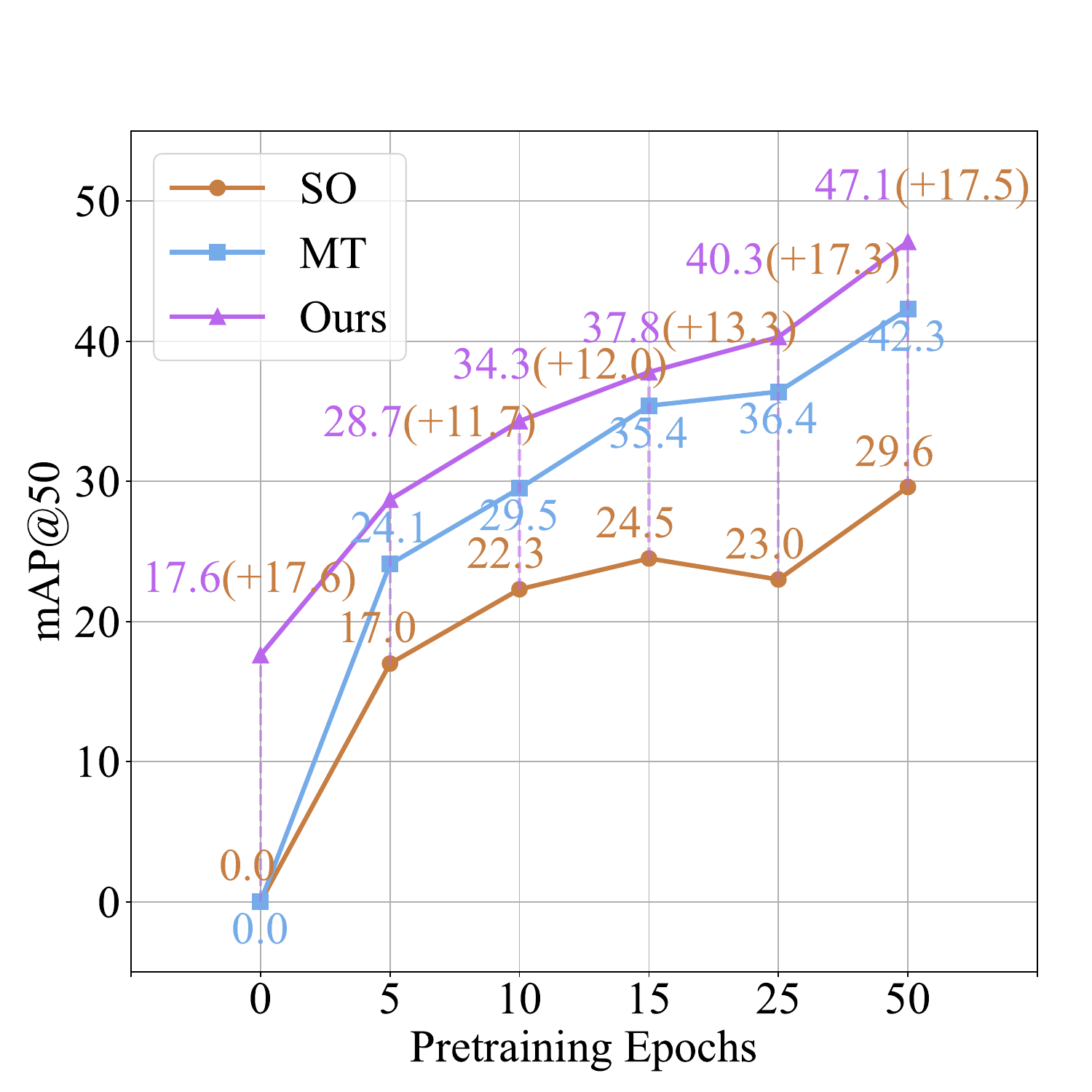}
\caption{Results of Source-only (SO), Mean Teacher (MT), and ours under restricted pretraining epochs and weak source pretraining performance.}
\label{fig:pretraining_epochs}
\end{center}
\end{figure}

\subsection{Zero-shot Grounding DINO Performance}
To provide a more comprehensive comparison, we further evaluate the zero-shot performance of Grounding DINO~\cite{liu2024grounding} on challenging cross-weather adaptation tasks, including Cityscapes-to-Foggy Cityscapes and Cityscapes-to-ACDC. We use Grounding DINO with the Swin-B backbone, identical to that used in DEPF. To ensure fairness given the backbone differences between Swin-B Grounding DINO and ResNet-50 Deformable DETR, we also compare against our method using Swin-B Deformable DETR. Results are reported in Tab.\ref{tab:backbone_full} and Tab.\ref{tab:acdc_full}.

Benefiting from large-scale pretraining, Grounding DINO achieves strong zero-shot results on diverse weather conditions (e.g., 34.3\% mAP on Foggy Cityscapes). However, by integrating the general visual knowledge from Grounding DINO with the domain-specific knowledge of the source detector, our approach consistently outperforms zero-shot Grounding DINO under the same backbone (e.g., 52.9\% vs. 34.3\% mAP on Foggy Cityscapes), demonstrating the effectiveness of our method and the advantage of leveraging VFMs for SFOD.

\subsection{VFM Backbone Selection} We investigate the effect of different Vision Foundation Model (VFM) backbones used in our PGFA and PIFA modules. As summarized in Tab.~\ref{tab:backbone_pseudo_label}, the choice of VFM significantly influences overall detection performance. Among all configurations, DINOv2 with a ViT-G/14 backbone achieves the highest mAP of 47.1\%, with performance constantly improving as the ViT model scale increases from ViT-S to ViT-G. This trend highlights the benefit of stronger semantic representations provided by larger vision transformers, which facilitate more accurate and robust cross-domain alignment.

Interestingly, although Grounding DINO is specifically designed for object detection and integrates multimodal supervision from vision-language data, its performance lags behind DINOv2-based variants—achieving only 46.2\% mAP at best with Swin-B. This performance gap may origin from the following reasons: (1) the comparatively smaller pretraining dataset used by Grounding DINO (1.82M images) versus DINOv2 (142M images)~\cite{oquab2023dinov2}, which may limit the diversity and generality of its learned features, and (2) the added complexity and potential domain mismatch introduced by vision-language feature spaces in Grounding DINO, which might be less optimal for purely visual alignment tasks like ours.

\begin{figure*}[t] 
  \centering
   \begin{overpic}[width=1.0\textwidth]{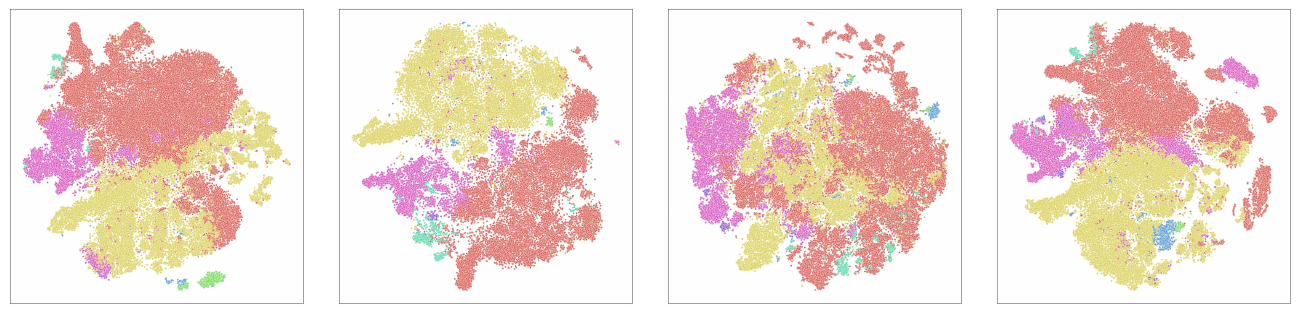} 
   \put(4,-1){(a) Source only}
   \put(30,-1){(b) Mean Teacher}
   \put(59,-1){(c) DRU}
   \put(84,-1){(d) Ours}
   \end{overpic}
  \caption{t-SNE visualization of instance-level features colored by predicted categories on the target domain. From left to right: (1) Source only, (2) Mean Teacher baseline, (3) DRU~\cite{khanh2024dynamic}, and (4) Ours. Each point represents an instance-level feature, and different colors indicate different predicted classes. Our method produces more compact and well-separated class clusters, indicating enhanced discriminability under domain shift.}
  \label{vis:tsne}
\end{figure*}

\begin{figure*}[t] 
  \centering
   \begin{overpic}[width=0.8\textwidth]{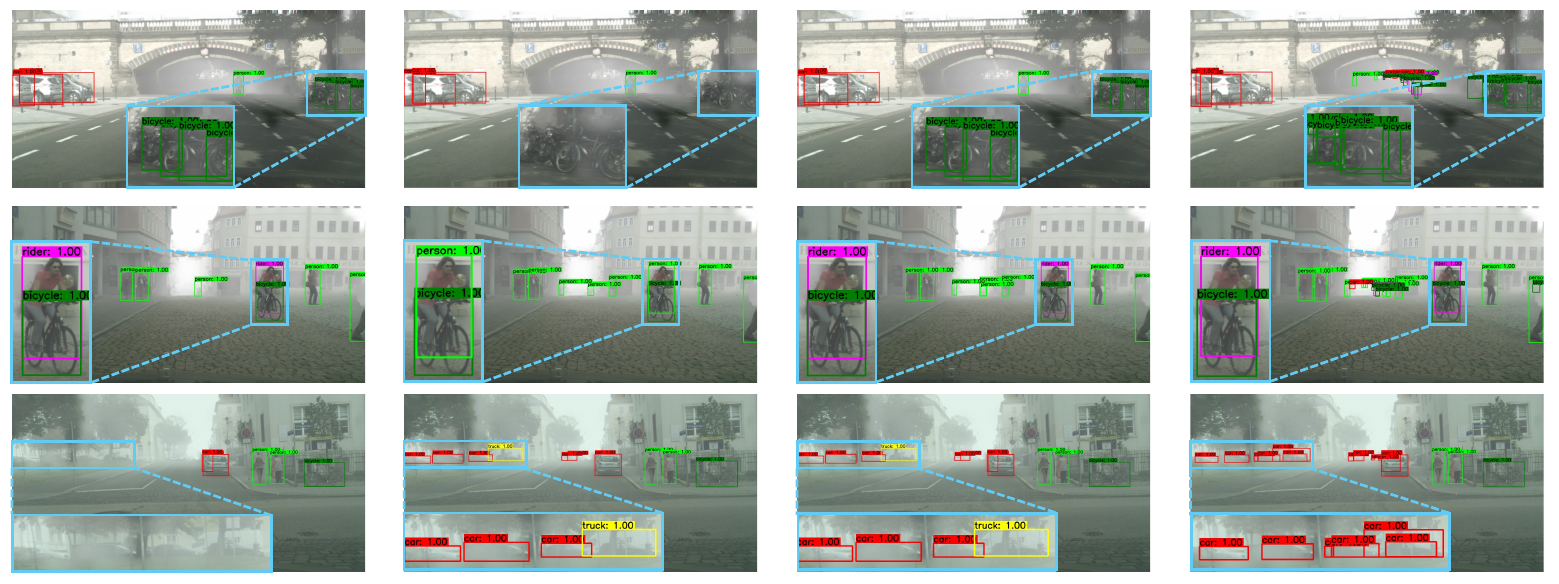} 
   \put(4,-1){(a) Teacher only}
   \put(29,-1){(b) GDINO only}
   \put(58,-1){(c) Ours}
   \put(79,-1){(d) Ground Truth}
   \end{overpic}
  \caption{Pseudo-label visualization on the Cityscapes-to-Foggy Cityscapes adaptation scenario among (a) Teacher only, (b) GroundingDINO (GDINO) only, (c) Ours (Teacher+GDINO fused), (d) Ground Truth. We zoom in the discriminative regions for each set of predictions.}
  \label{vis:pseudo_label}
\end{figure*}

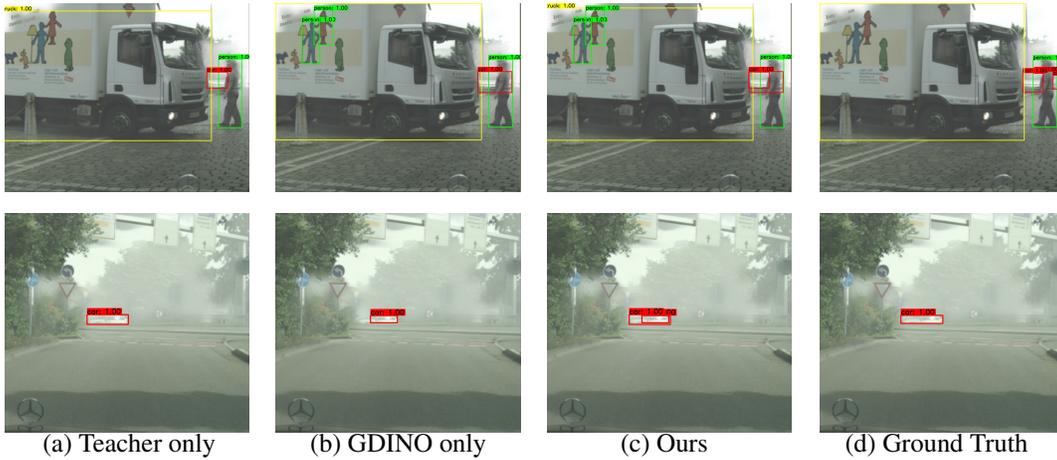
\begin{figure*}[t] 
  \centering
   \begin{overpic}[width=0.8\textwidth]{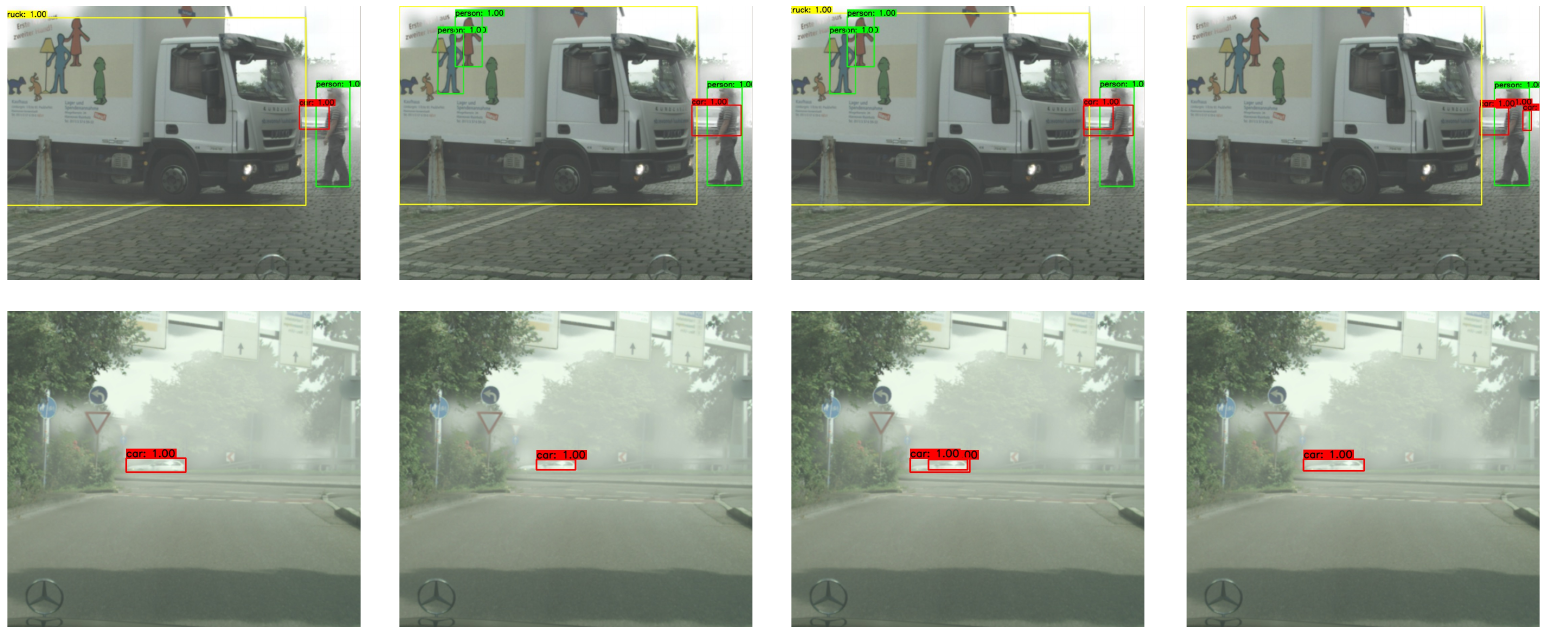} 
   \put(4,-1.4){(a) Teacher only}
   \put(29,-1.4){(b) GDINO only}
   \put(58,-1.4){(c) Ours}
   \put(79,-1.4){(d) Ground Truth}
   \end{overpic}
  \caption{Pseudo-label visualization (\textbf{failure cases}) on the Cityscapes-to-Foggy Cityscapes adaptation scenario among (a) Teacher only, (b) GroundingDINO (GDINO) only, (c) Ours (Teacher+GDINO fused), (d) Ground Truth. }
  \label{vis:failure_case}
\end{figure*}

\begin{table}[!t]
\centering\footnotesize
\caption{Performance comparison of different VFM backbone series (left) and pseudo-labeling strategies (right). ``RI'' means remove individual predictions.}
\begin{minipage}{0.49\linewidth}
\centering
\resizebox{\linewidth}{!}{%
\begin{tabular}{cc|c}
\toprule
Aligning VFM & Size & mAP \\
\midrule
\multirow{2}{*}{Grounding DINO} & Swin-S & 46.0 \\
                                & Swin-B & 46.2 \\
\midrule
\multirow{4}{*}{DINOv2} & ViT-S/14 & 46.2 \\
                        & ViT-B/14 & 46.7 \\
                        & ViT-L/14 & 46.8 \\
                        & ViT-G/14 & \textbf{47.1} \\
\bottomrule
\end{tabular}%
}
\end{minipage}
\hfill
\begin{minipage}{0.43\linewidth}
\centering
\resizebox{\linewidth}{!}{%
\begin{tabular}{lc|c}
\toprule
\multicolumn{2}{c|}{Pseudo labels} & mAP \\
\midrule
\multicolumn{2}{c|}{Teacher only} & 45.0 \\
\multicolumn{2}{c|}{GDINO only} & 43.3 \\
\midrule
\multirow{4}{*}{Fused} 
& NMS           & 46.3 \\
& WBF           & 46.7 \\
& ours + RI     & 43.2 \\
& ours w/o weight & 46.8 \\
& ours          & \textbf{47.1} \\
\bottomrule
\end{tabular}%
}
\end{minipage}
\label{tab:backbone_pseudo_label}
\end{table}

\begin{table}[!t]
\centering\footnotesize
\caption{Computational efficiency analysis. Comparison of training and inference time among the Mean Teacher baseline, PGFA \& PIFA (with DINOv2), and DEPF (with Grounding DINO).}
\small
\resizebox{\linewidth}{!}{
\begin{tabular}{ccc|ccc}
\toprule
Mean Teacher & PGFA \& PIFA & DEPF & mAP / Gain & $\Delta$Training time$\uparrow$ & $\Delta$Inference time$\uparrow$\\
\midrule
\Checkmark &         &       & 42.3 / - & +0.0\% & \multirow{4}{*}{+0.0\%} \\ 
\Checkmark & \Checkmark  &       & 45.0 / +2.7 & +28.6\% & \\
\Checkmark &   &  \Checkmark  & 45.9 / +3.6 & +42.8\% & \\
\Checkmark & \Checkmark & \Checkmark  & \textbf{47.1} / +\textbf{4.8} & +79.0\% & \\
\bottomrule
\end{tabular}
}
\label{tab:abl_efficiency}
\end{table}


\subsection{Pseudo-label Fusion Strategies} We compare multiple pseudo-label fusion strategies to validate DEPF. Tab.\ref{tab:backbone_pseudo_label} further reveals that relying solely on either the teacher model or Grounding DINO for pseudo-label generation results in suboptimal performance, underscoring the necessity of integrating predictions from both sources. Naively applying standard fusion methods like Non-Maximum Suppression (NMS) or Weighted Box Fusion (WBF)\cite{solovyev2021weighted} also falls short, which yields 0.8\% and 0.4\% lower mAP than our proposed method, respectively. This performance gap is likely due to two factors discussed in the main paper: the difficulty of tuning confidence weights and the negative impact of contradictory predictions that are not properly processed.

The importance of fusing contradictory predictions is further illustrated by the pseudo-label visualizations in Fig.~\ref{vis:pseudo_label}. In line 2 of the figure, the teacher and Grounding DINO assign different labels to the same object by predicting \texttt{rider} and \texttt{person}, respectively. If vanilla NMS or WBF is directly applied, both predictions may survive and be treated as valid pseudo-labels. This leads to label ambiguity and conflicts during training, ultimately harming optimization.

We also evaluate an alternative fusion strategy, Remove Individual (RI), which discards predictions not shared by both sources. While this approach avoids contradictory labels, it sacrifices recall and suppresses complementary information, resulting in inferior performance.

In contrast, our entropy-aware fusion strategy achieves the best performance (47.1\% mAP) by jointly refining both box locations and class logits based on uncertainty. This design not only resolves label conflicts but also leverages the strengths of both sources, leading to improved localization and classification quality.

\subsection{Computational Efficiency Analysis.} To provide a clearer understanding of the computational overhead introduced by each component, we analyze both training and inference efficiency, as summarized in Tab.~\ref{tab:abl_efficiency}. PGFA and PIFA are grouped together since they share the same feature extraction backbone from DINOv2, while DEPF relies on Grounding DINO for pseudo-label generation during inference. Compared with the Mean Teacher baseline, integrating PGFA \& PIFA increases training time by 28.6\% but brings a 2.7 mAP gain. Incorporating DEPF alone raises training cost by 42.8\% while achieving a 3.6 mAP improvement. When all modules are combined, the training overhead reaches 79.0\%, yet the overall performance gain is the highest (+4.8 mAP). Notably, the inference time remains unchanged since all VFMs are employed solely for offline feature extraction or pseudo-label generation, without introducing additional parameters or modules during inference. This design preserves deployment efficiency and ensures fair comparisons with baseline models.

\subsection{Robustness to Weak Source Pretraining} A key challenge in SFOD is \textbf{reliance on high-quality source-pretrained models}, which are often inaccessible due to privacy concerns. We simulate weak pretraining by limiting source model epochs, and we run our SFOD algorithm with the same training epochs and hyperparameters. Results in Fig.~\ref{fig:pretraining_epochs} show that even with minimal pretraining, our method still provides substantial gains, showing resilience to poor initialization. Notably, even without any source pretraining, our method still achieves 17.6\% mAP under the same number of adaptation epochs. Moreover, stronger source models result in larger performance gains (e.g., +11.7\% for 5 epochs, +17.5\% for 50 epochs), confirming that effectively leveraging limited source knowledge via VFMs is crucial and suggests a promising direction for future research.

\subsection{Failure Case Analysis}

Figure~\ref{vis:failure_case} presents representative pseudo-label failure cases in the Cityscapes-to-Foggy Cityscapes scenario. In the first row, our fusion mechanism fails to suppress a high-confidence false positive inherited from one source model. Although the complementary model correctly omits this region, the fusion retains the erroneous detection due to the dominant confidence weight, revealing a limitation in filtering model-specific false activations. In the second row, the fused pseudo-label exhibits inaccurate localization, where slight spatial misalignment between the teacher and Grounding DINO boxes leads to a blurred or misplaced fusion result. This indicates that when both sources provide coarse or inconsistent spatial cues, the fusion may propagate localization noise rather than refine it. These observations suggest that while the fusion strategy effectively enhances coverage in most cases, its robustness could be further improved through confidence calibration and geometry-aware alignment.

\subsection{t-SNE Visualization}
To further assess the discriminative capability of our method, we visualize the instance-level features on the target domain using t-SNE~\cite{maaten2008visualizing}, with color indicating the predicted class. As shown in Fig.~\ref{vis:tsne}, the source model (leftmost) exhibits severely entangled class distributions, reflecting its poor generalization to the target domain. The Mean Teacher baseline provides modest improvement, while DRU demonstrates stronger separation in some regions. In contrast, our method yields significantly more compact and clearly separated clusters, suggesting that the features learned through VFM-guided alignment and dual-source supervision are more discriminative and semantically coherent. These visual results align with our quantitative improvements and validate the effectiveness of our approach in enhancing class separability under domain shift.



\end{document}